\documentclass[runningheads]{llncs}

\usepackage{eccv}

\usepackage{eccvabbrv}

\usepackage{graphicx}
\usepackage{booktabs}

\usepackage[accsupp]{axessibility}  

\usepackage{hyperref}

\usepackage{orcidlink}

\definecolor{cvprblue}{rgb}{0.21,0.49,0.74}
\definecolor{darkgreen}{rgb}{0, 0.4, 0}
\definecolor{darkred}{rgb}{0.5, 0.1, 0.1}
\definecolor{darkblu}{rgb}{0, 0, 0.6}
\definecolor{bgblu}{rgb}{0.9, 0.96, 1}
\definecolor{pergreen}{rgb}{0, 0.7, 0}
\usepackage{amsmath}
\usepackage{multirow}
\usepackage{multicol}

\usepackage[ruled,vlined]{algorithm2e}
\usepackage{array,multirow,graphicx}
\usepackage{xr}
\usepackage[export]{adjustbox}

\newcommand{\dnp}{\texttt{DNP}\@\xspace}
\newcommand{\dns}{\texttt{DNS}\@\xspace}
\newcommand{\dna}{\texttt{auto-DNP}\@\xspace}
\newcommand{\dni}{$\bar{\mathcal{I}}$}
\newcommand{\supp}[1]{#1}

\begin{document}

\title{Improving image synthesis with diffusion-negative sampling} 


\author{Alakh Desai\inst{1} \and
Nuno Vasconcelos\inst{1}}

\authorrunning{A. Desai and N. Vasconcelos}

\institute{University of California San Diego, USA\\
\email{\{ahdesai,nuno\}@ucsd.edu}}

\maketitle

\begin{abstract}
    For image generation with diffusion models (DMs), a negative prompt $\bf n$ can be used to complement the text prompt $\bf p$, helping define properties not desired in the synthesized image. While this improves prompt adherence and image quality, finding good negative prompts is challenging. We argue that this is due to a semantic gap between humans and DMs, which makes good negative prompts for DMs appear unintuitive to humans. To bridge this gap, we propose a new {\it diffusion-negative prompting} (\dnp) strategy. \dnp is based on a new procedure to sample images that are least compliant with $\bf p$ under the distribution of the DM, denoted as {\it diffusion-negative sampling} (\dns). Given $\bf p$, one such image is sampled, which is then translated into natural language by the user or a captioning model, to produce the negative prompt $\bf n^*$. The pair (${\bf p}$, ${\bf n^*}$) is finally used to prompt the DM. \dns is straightforward to implement and requires no training. Experiments and human evaluations show that \dnp performs well both quantitatively and qualitatively and can be easily combined with several DM variants. 
    
\end{abstract}

\section{Introduction}\label{sec:intro}
\begin{figure}[t]
\centering 
\includegraphics[width=\linewidth]{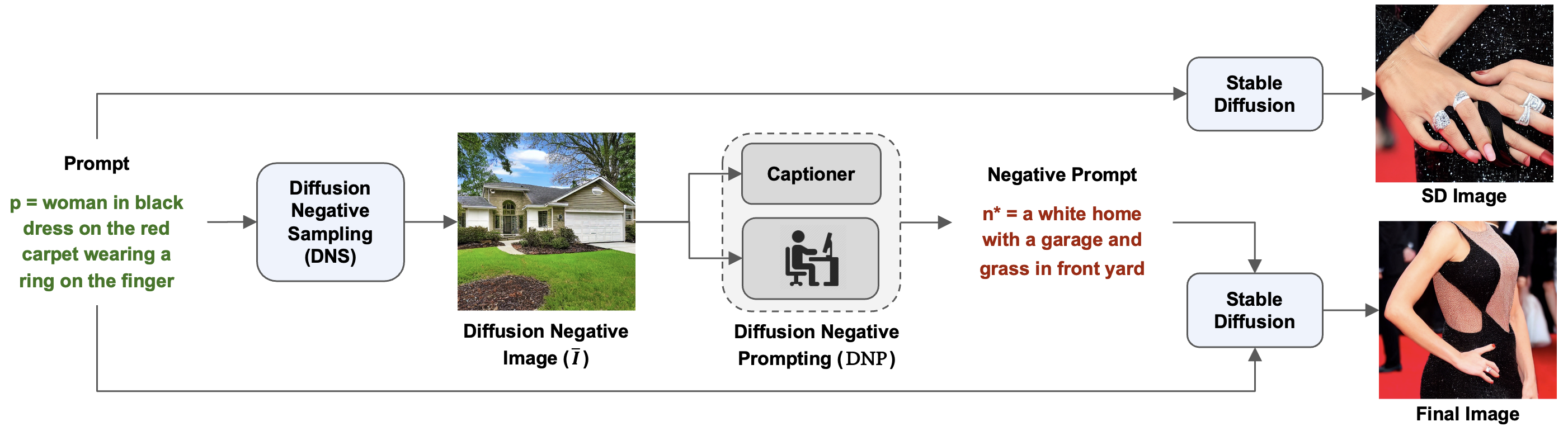}
\caption{\footnotesize \dnp improves quality of synthesis for prompts $\bf p$ ({\color{darkgreen}green}) of SD's images (top-right). A diffusion-negative image, \dni, is sampled using \dns, enabling the user to visualize the negation of $\bf p$ under DM's distribution. The user translates the \dni~into a negative prompt $\bf n^*$ ({\color{darkred}red}), by a process denoted as \dnp, and the DM is prompted with the pair $({\bf p},{\bf n^*})$. This increases compliance and quality of the synthesized image (bottom-right). Replacing the user with a captioning model is denoted as \dna.}
\label{fig:DNP}
\end{figure}

Diffusion models (DMs)~\cite{dalle2,ldm, imagen} have shown exquisite capacity to synthesize visually appealing images guided by textual prompts. However, they are not easy to control. While the synthesized images are usually impressive, various classes of prompts are known to be difficult, e.g. prompts involving humans~\cite{text2human}, hands~\cite{text2hand}, or multiple objects and their interactions~\cite{ae}. The synthesized image quality can often be unsatisfactory for such prompts, and prompt adherence is usually weak. This is illustrated in Figure~\ref{fig:DNP}, which shows an image synthesized by Stable Diffusion (SD) for a complex prompt $\bf p$. The problem can be mitigated by adding alternative conditioning inputs to the diffusion model, such as visual conditioning with sketches or layouts~\cite{controlnet,layoutdiff}. This, however, usually requires skilled users and can be labor-intensive. Ideally, it should be possible to improve consistency with text prompts alone. In this work, we explore negative prompting~\cite{compos}, which has been quite effective but is difficult to use. We argue that this difficulty stems from a {\it semantic gap\/} between the concept representations of a human user and the DM. We then introduce a new prompting technique, denoted as {\it diffusion-negative prompting} (\dnp), to bridge this gap, as illustrated in Figure~\ref{fig:DNP}.

\begin{figure*}[t!]
    \centering
    \begin{subfigure}[t]{0.43\textwidth}
        \resizebox{\linewidth}{!}{
            \begin{tabular}{ccc}
            \multicolumn{3}{c}{\color{darkgreen} ${\bf p}$: an airplane standing on the runway} \\
            \color{darkred}${\bf n}$: ``"& \color{darkred}${\bf n}$: flying & \color{darkred}${\bf n}$: soaring \\
            \includegraphics[width=0.6\linewidth]{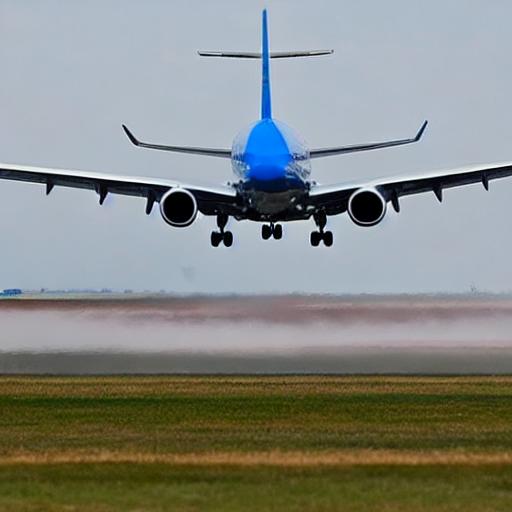}
            &\includegraphics[width=0.6\linewidth]{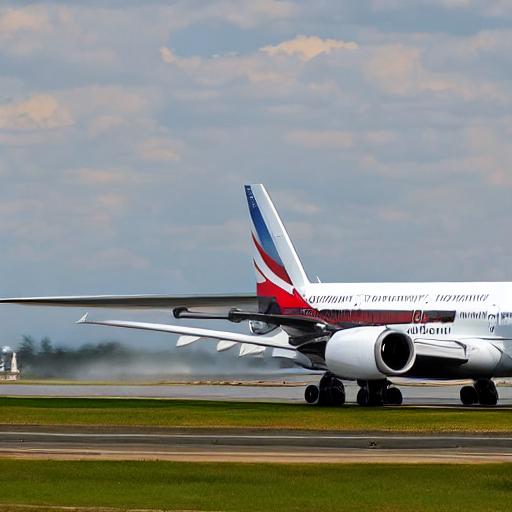}
            &\includegraphics[width=0.6\linewidth]{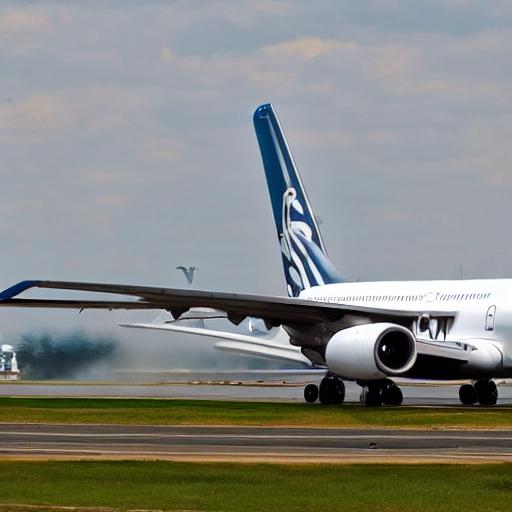}\\
            \multicolumn{3}{c}{\color{darkgreen} ${\bf p}$: an airplane standing on the runway} \\
            \color{darkred}${\bf n}$: ``"& \color{darkred}${\bf n}$: flying & \color{darkred}${\bf n}$: soaring \\
            \includegraphics[width=0.6\linewidth]{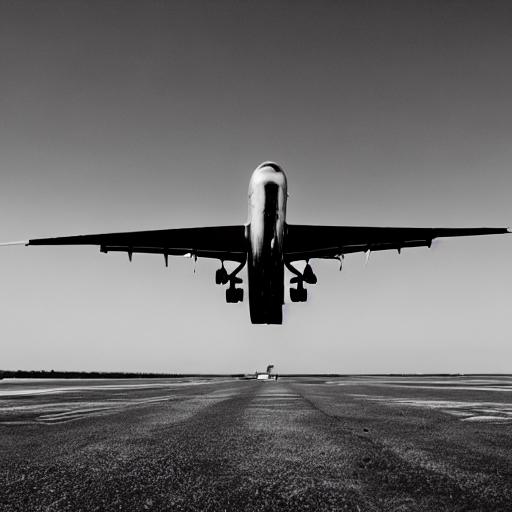}
            &\includegraphics[width=0.6\linewidth]{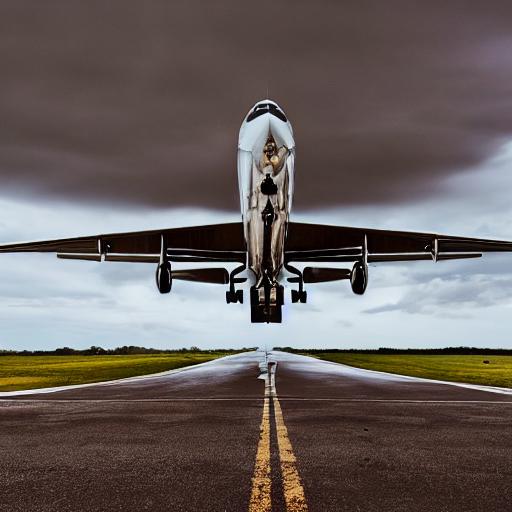}
            &\includegraphics[width=0.6\linewidth]{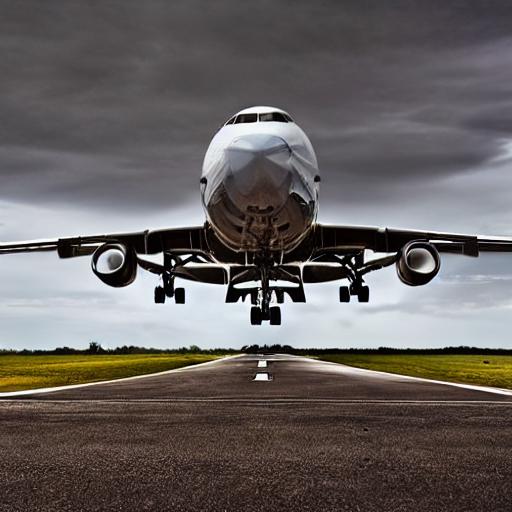}\\
            \end{tabular}}
            \caption{\textbf{Semantic negatives:} While effective, they do not guarantee compliance across seeds. Success (top), failure (bottom).}  
            \label{fig:semantic_neg_example}
    \end{subfigure}
    \hfill
    \begin{subfigure}[t]{0.56\textwidth}
        \resizebox{\linewidth}{!}{
            \begin{tabular}{@{}c@{\hskip 0.2em}c@{\hskip 0.2em}c@{\hskip 0.2em}c@{}}
            \multicolumn{4}{c}{\color{darkgreen} ${\bf p}$: a cat and a dog}\\
            \color{darkred}${\bf n}$: ``"& \color{darkred}${\bf n}$: two dogs& \color{darkred}${\bf n}$: two cats, two dogs & \color{darkred}${\bf n}$: ugly, brown \\
            \includegraphics[width=0.45\linewidth]{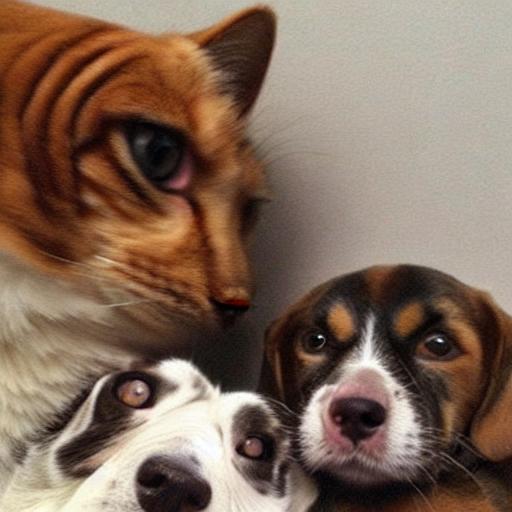}
            &\includegraphics[width=0.45\linewidth]{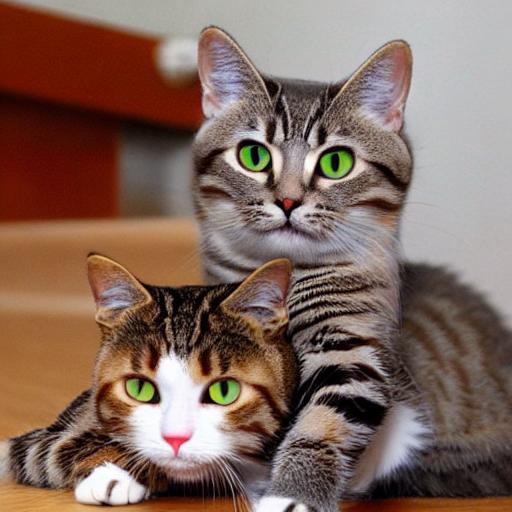}
            &\includegraphics[width=0.45\linewidth]{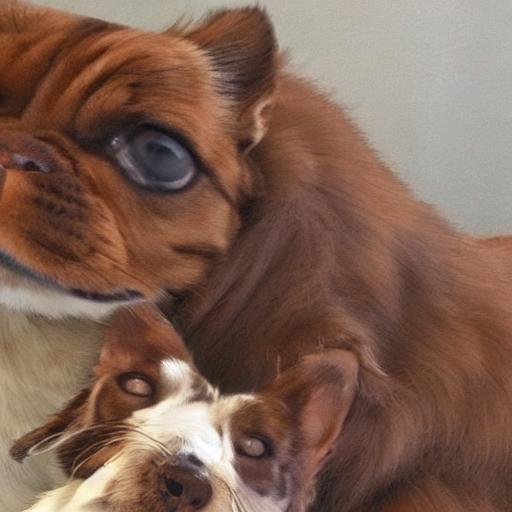}
            &\includegraphics[width=0.45\linewidth]{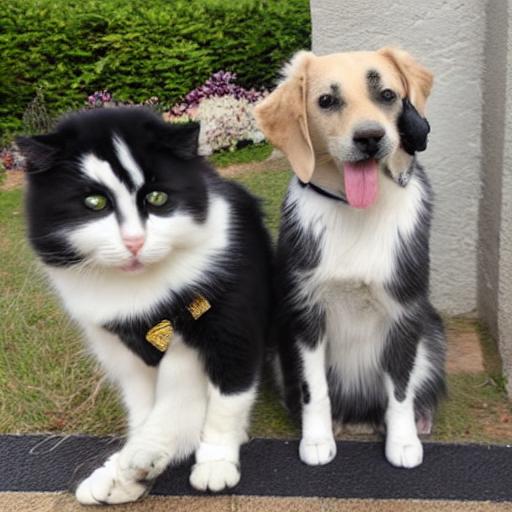}\\[0.05in]
            \multicolumn{4}{c}{\color{darkgreen} ${\bf p}$: a cat and a dog}\\
            \color{darkred}${\bf n}$: ``"& \color{darkred}${\bf n}$: two cats & \color{darkred}${\bf n}$: two cats, two dogs & \color{darkred}${\bf n}$: ugly, brown \\
            \includegraphics[width=0.45\linewidth]{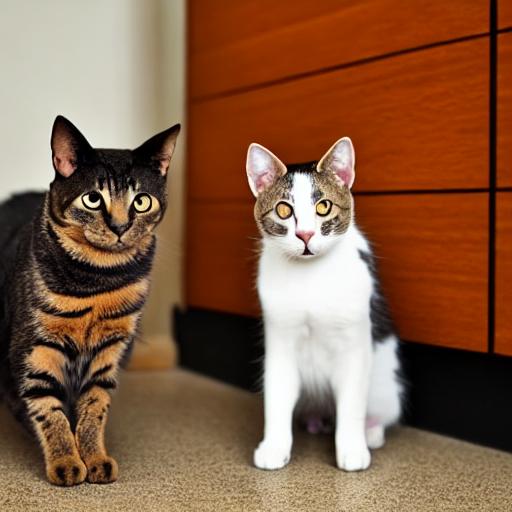}
            &\includegraphics[width=0.45\linewidth]{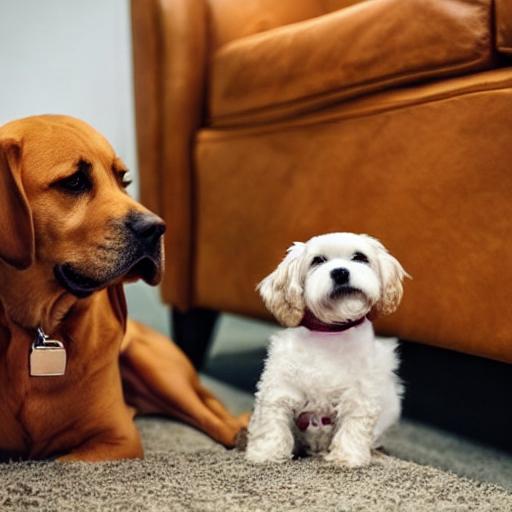}
            &\includegraphics[width=0.45\linewidth]{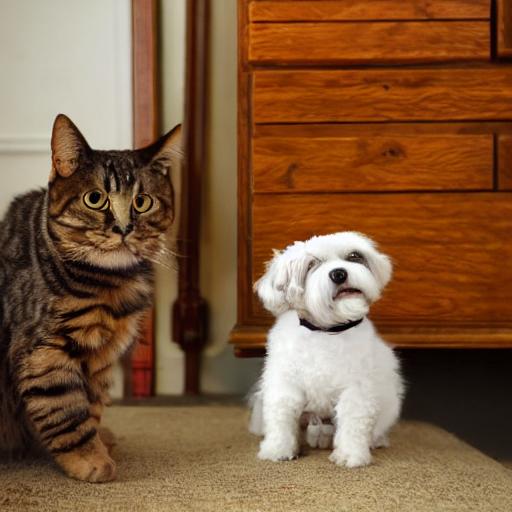}
            &\includegraphics[width=0.45\linewidth]{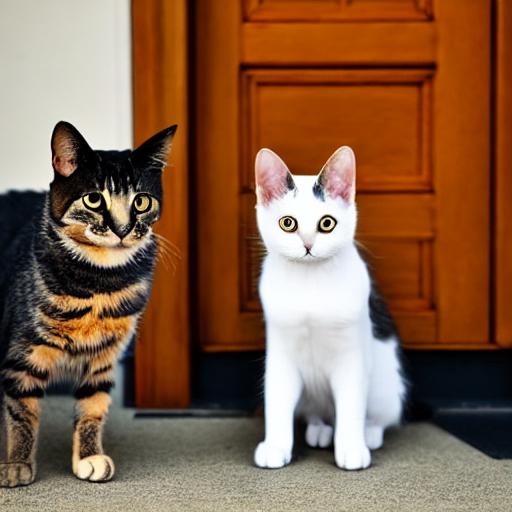}\\
            \end{tabular}}
            \caption{\textbf{Non-semantic negatives:} A good negative prompt needs multiple iterations for every seed as they can be highly unintuitive.}  
            \label{fig:nonsemantic_neg_example}
    \end{subfigure}
    \caption{\footnotesize{Example of negative prompting for both semantic and non-semantic scenarios. The positive prompt, ${\bf p}$ and negative prompts, ${\bf n}$ are on top of each image.}}
\end{figure*}

Negative prompting complements the text prompt ${\bf p}$, e.g. \textit{``an airplane standing on the runway"}, with negative prompts ${\bf n}$, e.g. \textit{``flying"} or \textit{``soaring"}. As illustrated in the top row of Figure~\ref{fig:semantic_neg_example}, this greatly improves compliance of the synthesized image with prompt ${\bf p}$ and can improve image quality. However, negative prompts are difficult to use, for two reasons. First, there are many prompts, e.g. see ${\bf p}$ = ``a cat and a dog" in Figure~\ref{fig:nonsemantic_neg_example}, without a clear negative. Second, even when an intuitive negative exists, e.g. ``flying" vs. ``standing", it is not necessarily a successful negative for DMs across seeds, as seen in the bottom row of Figure~\ref{fig:semantic_neg_example}. This raises the question: what is a good negative prompt for a DM? In general, {\it ${\bf n}$ is a good negative prompt if the images synthesized by prompt-pair $({\bf p},{\bf n})$ adhere more to $\bf p$ than those synthesized by ${\bf p}$ alone.\/} This, however, is a {\it DM-centric\/} definition of negative prompt. The problem is that the semantic representation of the DM is only a weak approximation to that used by a human. In general, DMs cannot replicate the human definition of concept negation. We refer to this problem as the {\it semantic gap\/} between DMs and humans. As a result, good negative prompts for a DM are frequently not intuitive for humans, i.e. what a human would consider a negative for $\bf p$. This makes it difficult for users to produce good negative prompts for DMs.

In this work, we address this problem by devising a strategy that enables humans to visualize the concepts that DMs consider negatives for prompt ${\bf p}$. This procedure is inspired by classifier-free guidance (CFG)~\cite{cfg}. While CFG uses a sampling guidance factor that increases the probability of images compliant with $\bf p$, we introduce a {\it diffusion-negative sampling\/} (\dns) guidance factor that encourages the sampling of the {\it diffusion-negative images\/}, \dni, least compliant with $\bf p$, under the DM's image distribution. This \dni~usually does not comply with the human understanding of the negation of concepts in $\bf p$. However, because it represents the DM's understanding of this negation, it can be shown to the user to overcome the semantic gap between them. The user can then produce a negative prompt $\bf n^*$ {\it in the language of the DM\/} by simply captioning \dni. Prompting the DM with the pair $({\bf p},{\bf n^*})$ usually produces better images than with $\bf p$ alone. 

Figure~\ref{fig:DNP} shows how \dnp can help when DMs produce poor-quality images. In response to the user prompt $\bf p$ = ``woman in black dress on the red carpet wearing a ring on the finger'', SD generates the top image, with distorted hands and rings. The rest of the figure illustrates \dnp. Given the prompt $\bf p$, the DM samples a negative image \dni$({\bf p})$ using \dns. In the figure, \dni~contains the house with a garden. The user inspects the image and produces a {\it diffusion-negative prompt} ${\bf n}^*$(\dni) = ``a white home with a garage and grass in the front yard'' that reflects the content of this image. This is similar to captioning the image, and can also be done automatically by an image captioning model, resulting in what we denote as \dna. Finally, the DM is prompted with the pair $({\bf p}, {\bf n^*})$ to produce an image that has much better compliance with $\bf p$ than the original. 

Our experiments show that \dnp improves prompt adherence and image quality, both in terms of quantitative metrics, like CLIP scores of image-prompt similarity, and subjective ratings by human evaluators. This is shown to hold both for vanilla DMs, like SD, and DMs optimized for solving specific problems, e.g. the \textit{A\&E}~\cite{ae} model tailored to synthesize images containing multiple objects. Various ablations also show that \dnp outperforms negative prompting solutions commonly used by practitioners, e.g. using a dictionary of universal negative prompts, random prompts, etc. It also provides much-needed transparency to the negation process. \dna provides a fully automated implementation of \dnp. While not as effective as \dnp, it maintains all the benefits discussed above, avoids user captioning effort, and allows automatic comparisons to other negative prompting procedures. Finally, our experiments consistently demonstrate the existence of a \textit{semantic gap} between DMs and humans. Although the negative images generated by \dns are frequently very unintuitive for humans, they help the DM significantly when translated into negative prompts.

Overall, this paper makes the following contributions. First, we hypothesize that DM image synthesis can be substantially enhanced by using negative prompts. However, these prompts are difficult for humans to generate, due to the \textit{semantic gap} between humans and DMs, which leads to a different understanding of concept negation. Second, we introduce the \dns procedure to overcome this \textit{semantic gap}, allowing humans to visualize the negative concept as understood by the DM. This enables negative prompting with \dnp, as shown in Figure~\ref{fig:DNP}. Third, we present extensive experiments on various datasets, showing that \dnp complements any existing DM or its variant. Fourth, we show that the process can be automated by using a pre-trained image captioner, leading to \dna. Finally, we show that \dnp is trivial to implement and requires no training.

\section{Diffusion Models} \label{sec:prelim}

\noindent\textbf{Latent Diffusion:} 
DMs combine a forward and a backward process for generative modeling. In the forward process, a sample is gradually corrupted by sequential application of small amounts of Gaussian noise. The backward process then sequentially denoises the sample using a learned neural network $\epsilon_\theta$. Latent DMs~(LDMs) increase the efficiency of the diffusion process by operating on the low dimensional latent space $\cal Z$ of an autoencoder, implemented with a pre-trained encoder/decoder pair $(\mathcal{E}(.), \mathcal{D}(.))$. This transforms image $x \in \mathcal{X}$ into latent representation $z = \mathcal{E}(x) \in {\cal Z}$ and reconstructs the image with $x = \mathcal{D}(\mathcal{E}(z))$.

\noindent\textbf{Training:} In the forward process, a noisy version of the latent representation is obtained at each time step $t$ with $z_t = \sqrt{\bar{\alpha}_t} z + \sqrt{1-\bar{\alpha}_t} \epsilon$. Here, $\bar{\alpha}_t = \prod_{s=1}^{t}(1-\beta_s)$, where $\{\beta_1,...,\beta_T\}$ is fixed according to the variance schedule and $\epsilon \sim \mathcal{N}(0, I)$. In the backward process, the denoising model $\epsilon_\theta$ estimates the noise $\epsilon_t$ added to the noisy latent representation at each time step $t$. The denoising is conditioned by the embedding $\tau({\bf p})$, of text prompt $\bf p$, where $\tau(.)$ is a text encoder. The denoising network parameters are learned by minimizing the loss
\begin{equation}
    \mathcal{L} = \mathcal{E}_{t\sim\mathcal{U}(1,T),\epsilon_t\sim\mathcal{N}(0,I)}\Bigl[  \lVert \epsilon_t - \epsilon_\theta(z_t;t,\tau({\bf p})) \rVert^2 \Bigr],
\end{equation}
where $\mathcal{U}$ and $\mathcal{N}$ are the uniform and Gaussian distributions, respectively.

\noindent\textbf{Inference:} Given prompt $\bf p$, images are sampled by iteratively alternating between denoising and sampling, with
\begin{equation}
    \hat{\epsilon}_p = \epsilon_\theta(z_t;t,\tau({\bf p})),\quad z_{t-1}=\textit{sample}(z_t,\hat{\epsilon}_p,t).
        \label{eq:hatep}
\end{equation}
The sampling method, \textit{sample}, varies with the DM. Popular approaches are DDPM~\cite{ddpm} and DDIM~\cite{ddim}. This process is initialized with a noise seed $z_T \sim \mathcal{N}(0, I)$ and produces a latent image code $z_0$, which is finally passed to the decoder to obtain image $x_0 = \mathcal{D}(z_0)$. 

\noindent\textbf{Classifier-free Guidance:}
CFG~\cite{cfg} is a method to trade off prompt compliance and image diversity. It consists of training the DM with and without prompt $\bf p$, randomly setting $\bf p$ to the empty prompt $\phi$="", with probability $P_{uncond}=0.1$. At inference, each denoising step uses a linear combination of the prompt-conditional $(\hat{\epsilon}_p)$ and unconditional  ($\hat{\epsilon}_\phi$) noise estimates 
\begin{equation}\label{eq:diff_inf}
    \hat{\epsilon} = \hat{\epsilon}_\phi + s(\hat{\epsilon}_p - \hat{\epsilon}_\phi),
\end{equation}
where $\hat{\epsilon}_p= \epsilon_\theta(z_t;t,\tau({\bf p}))$, $\hat{\epsilon}_\phi= \epsilon_\theta(z_t;t,\tau(\phi))$, and $s$ is an hyper-parameter that controls the conditioning strength, known as the guidance-scale. The modified noise $\hat{\epsilon}$ is then utilized to update the latent representation. 

\noindent\textbf{Negative Prompting:}
Despite conditioning on text prompt $\bf{p}$, prompt adherence of CFG can be weak when $\bf{p}$ refers to complex scenes, such as that in Figure~\ref{fig:DNP}. Negative prompting~\cite{compos} is useful as an extra conditioning input. Given negative prompt $\bf n$, the inference denoising steps are implemented  with 
\begin{equation}\label{eq:negp_emb}
    \hat{\epsilon} = \hat{\epsilon}_\phi + s(\hat{\epsilon}_p -\hat{\epsilon}_{n} )
\end{equation}
where 
\begin{equation}
    \hat{\epsilon}_{n} = \epsilon_\theta(z_t;t,\tau({\bf n})).
    \label{eq:hatepn}
\end{equation}

Practitioners have shown that simply replacing the empty prompt $\phi$ of CFG with the negative prompt $\bf n$, i.e. using $   \hat{\epsilon} = \hat{\epsilon}_{n} + s(\hat{\epsilon}_p -\hat{\epsilon}_{n}),$ yields similar results, albeit for a slightly different guidance scale $s$~\cite{AUTOMATIC1111}. This has the benefit of computational efficiency, as it only requires the calculation of two noise vectors instead of three. Hence, this method has gained popularity and has become the default implementation of negative prompting in many T2I generation models. However, in what follows, we use the theoretically more grounded (\ref{eq:negp_emb}) to derive a procedure to sample \textit{diffusion-negative} images (\dni).

\section{Creating Good Negative Prompts}
\noindent In this section, we introduce \dnp.
\subsection{Energy-based model interpretation}
Sampling with a DM of network $\epsilon_\theta(z_t;t,\tau({\bf p}))$ can be interpreted as sampling from probability distribution $p_\theta(z_t|{\bf p}) \propto e^{-E_\theta(z_t;t,\tau({\bf p}))},$ where $E_\theta$ is an energy function, using the score function $\nabla_z \log P(z|{\bf p})$ and Longevin dynamics~\cite{song2019generative, longevindyn}. The DM network is trained to learn the score function, i.e.
\begin{equation} \label{eq:DMscore}
    \epsilon_\theta(z_t;t,\tau({\bf p})) = -\nabla_{z_t} \log p_\theta(z_t|\tau({\bf p})).
\end{equation}
This motivates the CFG denoising step of~(\ref{eq:diff_inf}), which corresponds to sampling from the distribution
\begin{equation}
    \begin{aligned}
         p_{\textit{cfg}}(z_t, {\bf p}) \propto p_\theta(z_t) \gamma_{cfg}^s(z_t,{\bf p})
    \end{aligned}
\end{equation}
where
\begin{equation}
    \begin{aligned}\label{eq:gcfg}
         \gamma_{cfg}(z_t,{\bf p}) = \frac{p_\theta(z_t|{\bf p})}{p_\theta(z_t)}  \propto p_\theta({\bf p}| z_t)
    \end{aligned}
\end{equation}
is a guidance factor that increases the probability of codes $z_t$ corresponding to images compliant with prompt $\bf p$.
Similarly, the negative prompt denoising step of~(\ref{eq:negp_emb}) corresponds to sampling from 
\begin{equation}
    \begin{aligned}\label{eq:cfg_p}
         p_{\textit{np}}(z_t,{\bf p}, {\bf n}) \propto p_\theta(z_t) \gamma_{np}^s(z_t, {\bf p}, {\bf n})
    \end{aligned}
\end{equation}
with guidance factor
\begin{equation}
    \begin{aligned}\label{eq:cfg_np}
         \gamma_{np}(z_t,{\bf p}, {\bf n}) = \frac{p_\theta(z_t|{\bf p})}{p_\theta(z_t| {\bf n})}  \propto \frac{p_\theta({\bf p}| z_t)}{p_\theta( {\bf n}| z_t)}.
    \end{aligned}
\end{equation}
This increases the probability of sampling latent codes $z_t$ of large odds ratio
\begin{equation}
    o(z_t, {\bf p}, {\bf n}) = \frac{p_\theta({\bf p}| z_t)}{p_\theta( {\bf n}| z_t)}.
    \label{eq:oddr}
\end{equation}
Since the  Bayes decision rule for deciding between ${\bf p}$ and ${\bf n}$ is to choose ${\bf p}$ when $o(z_t, {\bf p}, {\bf n}) \ge 1$ and ${\bf n}$ otherwise, it increases the probability that sampled codes comply with prompt ${\bf p}$ and not with prompt ${\bf n}$.

\subsection{Limitations of Negative Prompting}

Besides being theoretically well-motivated, negative prompting frequently improves image quality, as illustrated in the top row of Figure~\ref{fig:semantic_neg_example}. However, negative prompts are difficult to specify, because a natural negative frequently does not exist or does not produce the intended results. Figure~\ref{fig:nonsemantic_neg_example} shows an example for prompt ${\bf p}$ = ``a cat and a dog''. Here, the user may need to attempt several negative prompts $\bf n$,  requiring multiple iterations of image synthesis, as illustrated in the figure. Eventually, the user notices the insistence of the DM in brown objects and uses ${\bf n}$ = \textit{``brown''}. This is complemented with \textit{``ugly''}, which practitioners have identified as a good general-use negative prompt. To compound the problem, a negative prompt successful for one random seed can be unsuccessful for another seed, i.e. good negative prompts depend on {\it both} prompt $\bf p$ and seed $z_T$, as illustrated in  Figure~\ref{fig:semantic_neg_example}. In summary, negative prompting is an art form and can be quite cumbersome, even when successful. 

In practice, the human-provided negative prompt frequently fails to improve the quality of the image generated by the DM. We hypothesize that this is because the semantic representation of the DM is only a weak approximation to that of humans. Hence, there is no guarantee that, under the probability distribution modeled by the DM, there will be a large classification margin between $\bf p$ and the human-provided $\bf n$. In this case, negative prompting fails to produce samples of large odds ratio $o(z_t, {\bf p}, {\bf n})$ and the prompt pair $({\bf p}, {\bf n})$ fails to induce the model to produce more prompt compliant images than those generated with CFG, i.e. the posterior probability  $p_\theta({\bf p}| z_t)$ of (\ref{eq:gcfg}). In the extreme case where $p_\theta({\bf p}| z_t) \approx p_\theta({\bf n}| z_t)$ for the samples of high  $p_\theta({\bf p}| z_t)$, 
$o(z_t, {\bf p}, {\bf n}) \approx 1$ for those samples and (\ref{eq:cfg_p}) looses all sensitivity to prompt $\bf p$.

\subsection{Diffusion-Negative Guidance}
The discussion above suggests that a good negative prompt {\it for the DM\/} ``grounds'' the odds-ratio of (\ref{eq:oddr}), providing a reference for the prompt $\bf p$, which is defined in terms of the margin to negative $\bf n$. However, this grounding must happen under the {\it internal\/} representation of the DM, which is not necessarily that of humans. The goal is to encourage samples more likely to comply with given prompt $\bf p$ than the negative prompt, under the conceptual representation {\it of the model\/}. We formalize this goal by defining the optimal negative prompt ${\bf n^*}$ {\it for the DM\/} as the one that maximizes the steepness of the odds ratio with $\bf p$, i.e.
\begin{equation}
    {\bf n}^*({\bf p}) = \arg\max_{\bf n} \nabla_{z_t} \log o(z_t, {\bf p}, {\bf n}).
\end{equation}
This is denoted as the \textit{diffusion-negative} prompt for $\bf p$. Using (\ref{eq:hatep}), (\ref{eq:hatepn}), and (\ref{eq:oddr}), 

\begin{equation}
    \nabla_{z_t} \log o(z_t, {\bf p}, {\bf n})=\nabla_{z_t} \log \frac{p_\theta({\bf p}| z_t)}{p_\theta( {\bf n}|z_t)} = \nabla_{z_t} \log \frac{p_\theta(z_t| {\bf p})}{p_\theta(z_t| {\bf n})} = \hat{\epsilon}_p -\hat{\epsilon}_{n}. \label{eq:nabla}
\end{equation}

Hence, for a given $z_t$, a natural alternative definition of the \textit{diffusion-negative prompt} is that which induces the noise vector maximally distant from $\hat{\epsilon}_p$, i.e.
\begin{equation}
    {\epsilon}_{n^*} = \underset{{n| \left\|\hat{\epsilon}_{n}\right\|^2 = K}}{\arg\max} \left\|\hat{\epsilon}_p -\hat{\epsilon}_{n}\right\|^2
    \label{eq:ndopt}
\end{equation}
where $K$ prevents the noise magnitude from becoming infinitely large. This is the prompt that induces the steeper log odds surface at $z_t$, encouraging the largest margin between the probabilities $p_\theta(z_t|{\bf p})$ and $p_\theta(z_t|{\bf n}^*)$ in the neighborhood of $z_t$.  In \supp{Supplementary Section 1}, we show that (\ref{eq:ndopt}) has a solution
\begin{equation}\label{eq:nd}
    \hat{\epsilon}_{n^*} = -\sqrt{K} \frac{\hat{\epsilon}_p}{\left\|\hat{\epsilon}_p\right\|} \propto - \hat{\epsilon}_p.
\end{equation}
Choosing $K = \left\|\hat{\epsilon}_p\right\|^2$, i.e. equal noise strength under the positive and negative conditions, results in $\hat{\epsilon}_{n^*} = - \hat{\epsilon}_p$.

\subsection{Diffusion-Negative Sampling (\dns)}\label{sec:sample_neg_img}
It follows from~(\ref{eq:DMscore}) that 
\begin{equation} \label{eq:DMscoreNeg}
   \nabla_{z_t} \log p_\theta(z_t|{\bf n}^*)   = -\nabla_{z_t} \log p_\theta(z_t|{\bf p}) \implies p_\theta(z_t|{\bf n}^*)   \propto \frac{1}{p_\theta(z_t|{\bf p})}.
\end{equation}

However, for most  $p_\theta(z_t|{\bf n})$, (\ref{eq:DMscoreNeg}) is an improper distribution~\cite{improperdist}. This stems from the fact that a negative condition {\em must} be specified in context~\cite{compos_prev}. A general contextual constraint for DM-based sampling is that the negatively conditioned DM should still sample images from the image distribution $p_\theta(z_t)$. To account for this, we define the \textit{diffusion-negative} guidance factor as
\begin{equation} \label{eq:gnd}
    \gamma_{dn}(z_t,{\bf p}) = \frac{p_\theta(z_t)}{p_\theta(z_t|{\bf p})} \propto \frac{1}{p_\theta({\bf p}|z_t)}, 
\end{equation}
which leads to the \textit{diffusion-negative} distribution
\begin{equation}
    \begin{aligned}\label{eq:cfg_neg}
         p_{\textit{dn}}(z_t,{\bf p}) &\propto p_\theta(z_t) \gamma_{dn}^s(z_t,{\bf p})  
    \end{aligned}
\end{equation}
and inference denoising equation
\begin{equation}\label{eq:diff_inf_neg}
    \hat{\epsilon} = \hat{\epsilon}_\phi +s(\hat{\epsilon}_\phi - \hat{\epsilon}_p).
\end{equation}

Sampling with this equation is denoted as {\it diffusion-negative sampling\/} (\dns). When compared to (\ref{eq:cfg_np}), the \textit{diffusion-negative} guidance factor of (\ref{eq:gnd}) has two main differences. First, it does not require the semantic negative prompt $\bf n$, reflecting the fact that the definition of negative is fully based on the internal representation of the DM. Second, it replaces the odds ratio of (\ref{eq:oddr}) by the ratio between the image distribution $p_\theta(z_t)$ and the image distribution under the positive condition ${\bf p}$. This emphasizes sampling images {\it from the DM probability distribution with the lowest prompt conditional probability\/} $p_\theta(z_t|{\bf p})$. 
When compared to the guidance factor~(\ref{eq:gcfg}) of CFG,  (\ref{eq:gnd}) simply flips the role of the image distributions  $p_\theta(z_t)$ and $p_\theta(z_t|{\bf p})$. This provides a simple interpretation of \textit{diffusion-negative} guidance as the ``opposite'' of CFG. Rather than sampling natural images that align best with the condition $\bf p$, it emphasizes sampling natural images that least comply with it. 

The relation between the guidance factors of (\ref{eq:gnd}) and (\ref{eq:cfg_np})
\begin{equation}
    \gamma_{dn}(z_t,{\bf p}) = \frac{p_\theta(z_t)}{p_\theta(z_t|{\bf p})}  = \gamma_{np}(z_t,\phi, {\bf p}) 
\end{equation}
makes the implementation of \dns trivial for any model that supports negative prompting. It suffices to prompt the latter with empty positive prompt  $\phi$ and negative prompt ${\bf p}$. Hence, \dns~{\it requires no retraining of the DM, nor any optimizations at inference.\/}

\subsection{Diffusion-Negative Prompting (\dnp)}

\dnp relies on \dns to implement the procedure of Figure~\ref{fig:DNP}. Given prompt $\bf p$, \dns is used to sample a \textit{diffusion-negative} image, \dni, as discussed above. This image is then captioned by the user, to produce the \dnp ${\bf n^*}$. The DM is finally prompted with the prompt pair $({\bf p}, {\bf n^*})$ to produce the final image. As illustrated in Figure~\ref{fig:DNP}, the captioning step can be performed by either the user or any captioning model. In the latter case, the process is denoted \dna. \dna has the advantage of fully automating \dnp, making the entire process transparent to the user and reducing user effort. The price is a small increase in the failure rate due to captioning errors. In this paper, we use BLIP2~\cite{blip2} as the captioning model to implement \dna. However, \dna can be used with any other captioner. We show results obtained with GPTv4~(\textit{gpt-4-vision-preview}) in \supp{Supplementary Section 4}. Figure~\ref{fig:teaser} and the final column of Figure~\ref{fig:meth_exp} illustrate how \dnp produces images that either have higher quality or comply better with the prompt $\bf p$.

\section{Experiments}
In this section, we discuss the experimental evaluation of \dnp. 
\subsection{Datasets}
For a comprehensive dataset description see \supp{Supplementary Section 2.2}

\noindent\textbf{Existence and Attribute Binding~(A\&E) Dataset:}
This is the benchmark dataset introduced by \cite{ae}. This dataset is focused on entity neglect, which occurs when one or more entities are completely ignored by the generative model, and attribute assignment, which evaluates whether an attribute is associated with the correct object in the image. To evaluate these, each prompt in the dataset comprises two entities and their associated attributes. There are three categories of prompts: (1) “an {animal} and an {animal}”; (2) “a {color} {in-animate object} and an {animal}”; (3) “a {color} {in-animate object} and a {color} {in-animate object}”. We sample 50 prompts from each category for 32 random seeds. 

\noindent\textbf{Human and Hand Generation~(H\&H) Dataset:}
Humans and hands occur in various shapes and forms, including gender, race, activity, and individual differences. Despite the size of the training datasets, this variation has posed a major challenge to DMs, which are known to synthesize images with extra or merged limbs and fingers. We curated a challenging benchmark dataset to test the effect of \dnp on human and hand generation. The dataset consists of $45$ human-based prompts such as ``{a man wearing a sombrero}'' and $25$ hand-based prompts such as ``{hand with a ring on it}''. We run all prompts for $32$ seeds.

\subsection{Evaluation Metrics} \label{sec:metrics}

\noindent\textbf{Text-Image CLIP Score:}~\cite{clipscore} is a widely used metric of the similarity between a generated image and its text prompt. For the H\&H dataset, we calculate the CLIP Score between each generated image and its corresponding prompt. For the A\&E dataset, we follow the evaluation protocols established by \cite{ae}, which measure CLIP Score at full prompt and entity level. For evaluating the \textit{Full Prompt} CLIP Score, we calculate the score between the generated image and the full prompt. For the \textit{Minimum Object} CLIP Score, we take the minimum score between the entities in the prompt, as the entity with the minimum CLIP Score represents the neglected entity for the given prompt. This allows us to measure the CLIP Score at two levels, thereby providing better insight into the presence of both entities in the A\&E dataset prompts.

\noindent\textbf{Inception Score~(IS):}~\cite{IS} is a popular metric for assessing the quality and realism of images. It is the exponential of the average of the entropy of the label distribution predicted by the Inception~v3 classifier model. In particular, we choose IS over FID as it does not require a dataset with real images. 

\noindent\textbf{Human Evaluation:}
While CLIP Score and IS provide an objective measure of correctness and quality, they are known for not being fully aligned with human aesthetics and preferences. To evaluate \dnp in terms of human preferences, we used Amazon Mechanical Turk~(AMT) to evaluate a subset of each dataset. We performed human evaluation on $100$ prompt-seed pairs for the human dataset, $100$ pairs for the Hands dataset, and $150$~($50$ per category) pairs for the A\&E dataset. The evaluators were asked to compare the images using two criteria. 1) Adherence to the prompt, and 2) Quality (natural or realistic nature) of the image. MTurkers were tasked with choosing one image based the two criteria. They could also choose \textit{no clear winner} if they did not prefer one over the other. More details can be found in the \supp{Supplementary Section 2.3}.  

\subsection{Benchmarking \dnp vs \dna}

The evaluation of \dnp requires human users. PhD students, with no experience in visual language research, were shown an image and asked to produce a caption shorter than 60 words. Due to the prohibitive resources required to caption the image \dni~for all prompt-seed pairs of all the datasets, we restricted the evaluation of \dnp to the prompt-seed pairs used for human evaluation. We compared the performance of \dnp and \dna on this set and used \dna in the remaining experiments. Besides being cost-effective, this enables the replication of our experiments for comparison to other negative prompting methods. 
Table~\ref{tab:dnpvdna} compares the performance of \dnp and \dna, on the prompt-seed pairs used for the human evaluation\footnote{IS score cannot be computed in this case as the number of seeds per prompt is not large enough for IS to provide reliable results.}. For the rest of the paper, we use the notation DM+X, where DM is the diffusion model and X the negative prompting method.

While \dnp achieves the best performance, \dna provides almost identical results. This suggests that the BLIP2 model is quite effective at captioning the diffusion-negative negative images \dni.

\begin{table*}[t]
\centering
\resizebox{0.7\linewidth}{!}{
\begin{tabular}{c@{\hspace{1em}}c@{\hspace{1em}}c@{\hspace{1em}}c@{\hspace{1em}}c}
\hline
\multirow{3}{*}{\textbf{Method}} & \multicolumn{4}{c}{\textbf{CLIP Score}}\\
\cline{2-5}
 & \multirow{2}{*}{\textbf{Human}} & \multirow{2}{*}{\textbf{Hand}} & \multicolumn{2}{c}{\textbf{A\&E}} \\
\cline{4-5}
  &  & & \textbf{Min. Object} & \textbf{Full Prompt} \\
 \hline \hline
SD+\dna & 0.330 & 0.323 & 0.250 & 0.344 \\
SD+\dnp & \textbf{0.331}& \textbf{0.324} & \textbf{0.253} &\textbf{0.345} \\
\hline
\end{tabular}}
\caption{\footnotesize{Benchmarking \dnp to \dna with CLIP Score.}}
\label{tab:dnpvdna}
\end{table*}

\begin{table}
    \begin{subtable}{\linewidth}
        \centering
        \resizebox{0.8\linewidth}{!}{
        \begin{tabular}{cccccc}
        \hline
        \multirow{2}{*}{\textbf{Method}} & \multicolumn{2}{c}{\textbf{CLIP Score}} & \multirow{2}{*}{\textbf{IS}} & \multicolumn{2}{c}{\textbf{Human Evaluation}} \\
        \cline{2-3}\cline{5-6}
         & \textbf{Min. Object} & \textbf{Full Prompt} &  & \textbf{Correctness} & \textbf{Quality} \\
         \hline \hline
         No Clear Winner & - & - & - & 16.88\% & 10.46\% \\
        Stable Diffusion~(SD) & 0.242 & 0.335 & 13.17 & 18.90\% & 25.73\% \\
        SD+\dna & \textbf{0.258}~({\color{pergreen} +6.61\%}) & \textbf{0.346}~({\color{pergreen} +3.28\%}) & \textbf{13.35}~({\color{pergreen} +1.37\%}) & \textbf{64.22}\% & \textbf{63.81}\% \\
        
        \hline
        No Clear Winner & - & - & - & 16.95\% & 7.35\% \\
        Attend \& Excite~(\textit{A\&E}) & 0.264 & 0.349 & 11.86 & 30.06\% & 33.12\% \\
        \textit{A\&E}+\dna & \textbf{0.276}~({\color{pergreen} +4.54\%}) & \textbf{0.362}~({\color{pergreen} +3.72\%})  & \textbf{13.12}~({\color{pergreen} +10.62\%}) & \textbf{52.99}\% & \textbf{59.53}\% \\
        
        \hline
        \end{tabular}}
        \caption{\footnotesize{Quantitative Results on the A\&E dataset.}}
        \label{tab:ae}
    \end{subtable}
    \centering
    \begin{subtable}{\textwidth}
        \centering
        \resizebox{0.8\linewidth}{!}{
        \begin{tabular}{cccccc}
        \hline
        \multirow{2}{*}{\textbf{Dataset}} & \multirow{2}{*}{\textbf{Method}} & \multirow{2}{*}{\textbf{CLIP Score}} & \multirow{2}{*}{\textbf{IS}} & \multicolumn{2}{c}{\textbf{Human Evaluation}} \\
        \cline{5-6}
         & &  &  & \textbf{Correctness} & \textbf{Quality} \\
         \hline \hline
         \multirow{3}{*}{\textbf{Human Prompts}}&No Clear Winner & - & - & 14.42\% & 5.10\% \\
         &Stable Diffusion~(SD) &  0.322 & 9.95 & 27.60\% & 29.0\%\\
        &SD+\dna & \textbf{0.331}~({\color{pergreen} +2.80\%})& \textbf{10.13}~({\color{pergreen} +1.80\%}) & \textbf{57.98\%} & \textbf{65.80\%} \\
        
        \hline
         \multirow{3}{*}{\textbf{Hand Prompts}}&No Clear Winner & - & - & 19.89\% & 8.18\% \\
         &Stable Diffusion~(SD) & 0.309 & 12.04 & 20.45\% & 22.47\%\\
        &SD+\dna & \textbf{0.321}~({\color{pergreen} +3.88\%})) & \textbf{12.39}~({\color{pergreen} +2.94\%}) & \textbf{59.66\%} & \textbf{69.35\%} \\
        
        \hline
        \end{tabular}}
        \caption{\footnotesize{Quantitative Results on the H\&H dataset.}}
        \label{tab:hh}
    \end{subtable}
    \caption{\footnotesize{Quantitative Results: comparing prompt adherence and image quality by CLIP Score, IS and Human Evaluation. The percentage gains are shown in ({\color{pergreen} brackets})}}
\end{table}

\subsection{Quantitative Results}

We next discuss the results of quantitative experiments on image quality using \dna. Various ablations are also presented in \supp{Supplementary Section 4}.

\noindent\textbf{A\&E dataset:} We first compare  SD to SD+\dna, on the A\&E dataset in Table~\ref{tab:ae}. SD+\dna achieves a \textit{Min. Object} CLIP Score of $0.258$, which is 6.6\% higher than that of SD, illustrating that SD+\dna generates images that align better with the prompts. Since CLIP scores practically range from [0, 0.4]~\cite{clipscore}, the full prompt CLIP Score of $0.346$ of SD+\dna suggests good alignment between prompt and image. Human evaluation underscores the true strength of \dnp, as human evaluators prefer images generated by SD+\dna $3.4\times$ more in terms of correctness and $2.5\times$ more in terms of quality than those generated by SD. Note that SD+\dna even outperforms the \textit{A\&E} model in IS and matches its CLIP Score for both full prompt and minimum object, despite not being specifically designed for multiple object generation.

To demonstrate the flexibility of the \dnp approach, we also applied it to the \textit{A\&E} model, comparing \textit{A\&E}+\dna to \textit{A\&E}. We observe a moderate increase in the CLIP Score since \textit{A\&E} is already quite efficient at generating multiple objects. \textit{A\&E}+\dna achieves a \textit{Min. Object} CLIP Score of $0.276$, which is very close to the theoretical \textit{upper bound} for \textit{Min. Object} CLIP Score of $0.29$ as defined by \cite{ae}. The 10.62\% improvement in IS shows that \textit{A\&E} is more inclined to generate unrealistic images and the addition of \dnp improves its realism. The human evaluation also reflects this, as around $53\%$ of the evaluators prefer \textit{A\&E}+\dna over the \textit{A\&E} images for correctness and $60\%$ for quality, which is close to $1.8\times$ the preference for \textit{A\&E}.

\noindent\textbf{H\&H dataset:}  Table~\ref{tab:hh} summarizes the ability of \dnp to address the specific SD weaknesses of synthesizing humans and hands. Since both CLIP and IS may ignore distorted or maligned results when considering correctness or realism, they fail to give substantial insight into image quality (despite improvement). The human evaluation is much more important for this task. SD+\dna outperforms SD by a substantial margin, as evaluators prefer SD+\dna $~3\times$ more for correctness and $~2\times$ more for quality, for both human and hand prompts. Since evaluators were specifically tasked with checking the number and pose of limbs and fingers and the quality of the faces, their preference for images generated with \dnp shows that the latter improves SD performance substantially.

\begin{figure}[t]
\centering
\resizebox{\linewidth}{!}{
\begin{tabular}{c@{\hskip 0.05em}cc@{\hskip 0.05em}cc@{\hskip 0.05em}c}
\multicolumn{2}{c}{\Large \textbf{Prompts with Humans}} & \multicolumn{2}{c}{\Large \textbf{Prompts with Multiple Nouns}} & \multicolumn{2}{c}{ \Large \textbf{Prompts with Hands}}\\[0.1in]

\includegraphics[width=.34\linewidth,valign=m]{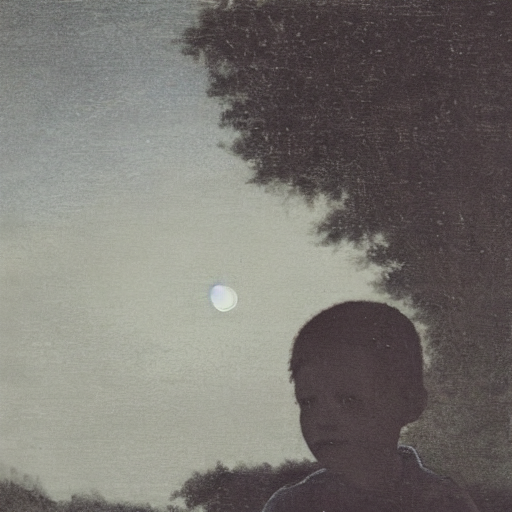} & \includegraphics[width=.34\linewidth,valign=m]{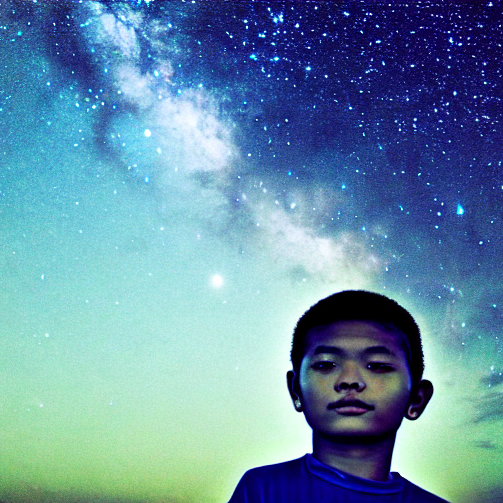} & \includegraphics[width=.34\linewidth,valign=m]{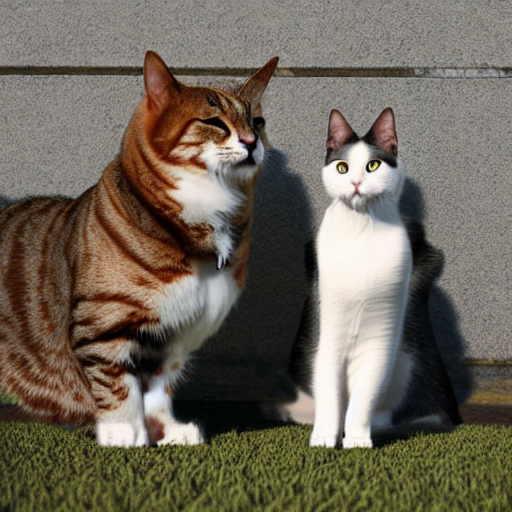} &
\includegraphics[width=.34\linewidth,valign=m]{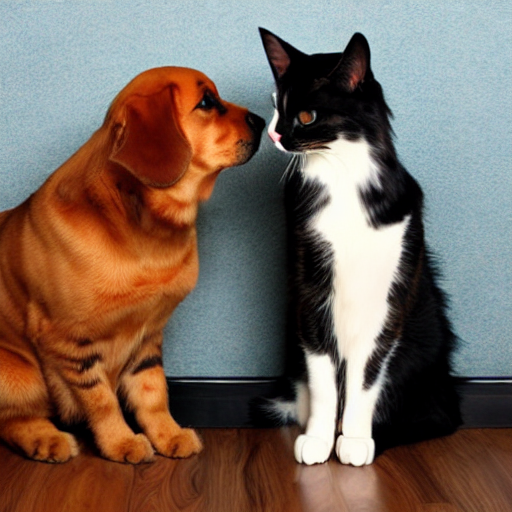} & \includegraphics[width=.34\linewidth,valign=m]{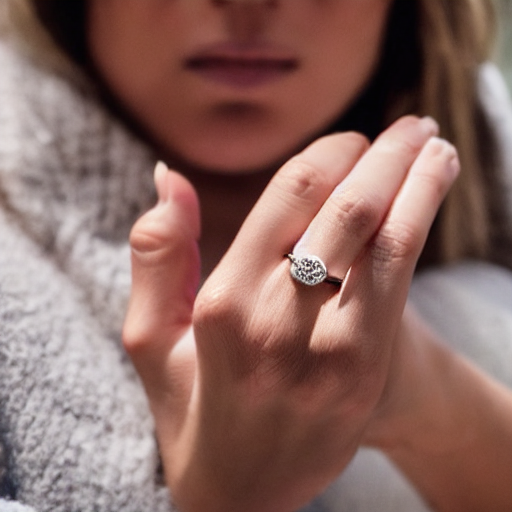} & \includegraphics[width=.34\linewidth,valign=m]{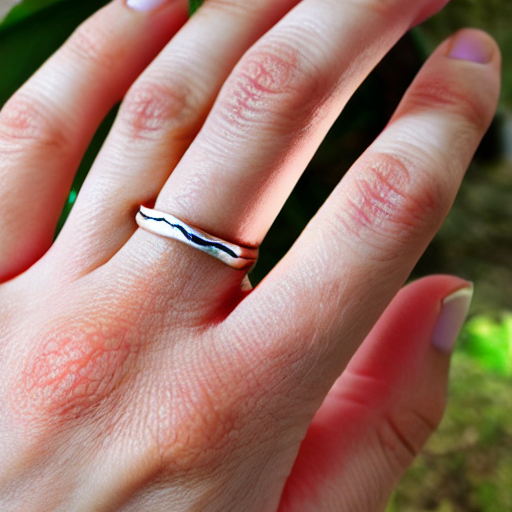}\\
\multicolumn{2}{c}{ \color{darkgreen} ${\bf p}$: a boy in the moonlight} & \multicolumn{2}{c}{\color{darkgreen}  ${\bf p}$: a cat and a dog} & \multicolumn{2}{c}{ \color{darkgreen} ${\bf p}$: a hand with a ring on it}\\
\multicolumn{2}{c}{ \color{darkred} ${\bf n^*}$: a living room with a fireplace and wooden beams} & \multicolumn{2}{c}{ \color{darkred} ${\bf n^*}$: an aerial view of a resort with a swimming pool} & \multicolumn{2}{c}{ \color{darkred} ${\bf n^*}$: a woman in a black  dress on the red carpet}\\[0.1in]

\includegraphics[width=.34\linewidth,valign=m]{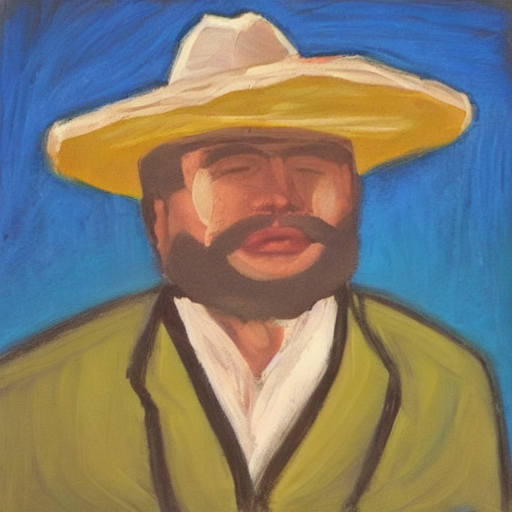} & \includegraphics[width=.34\linewidth,valign=m]{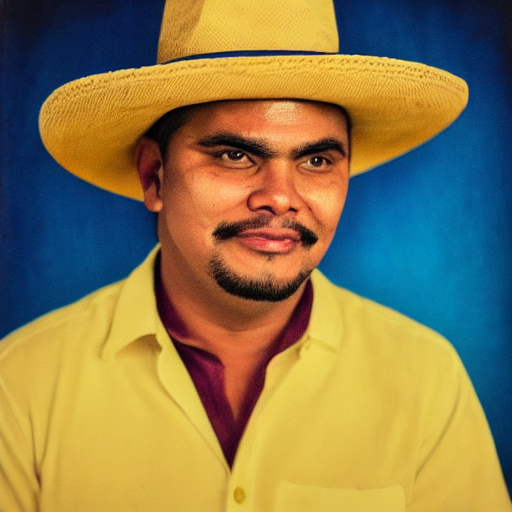} & \includegraphics[width=.34\linewidth,valign=m]{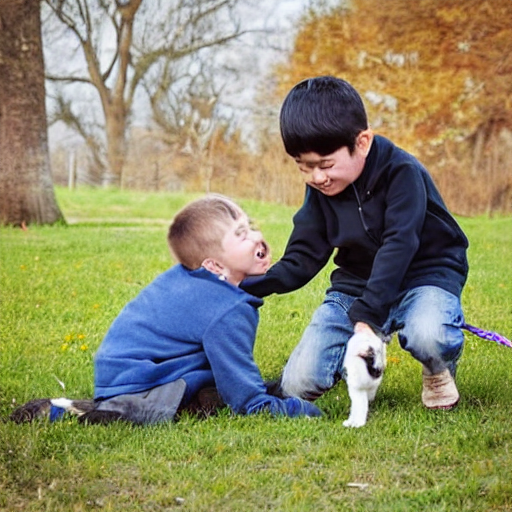} &
\includegraphics[width=.34\linewidth,valign=m]{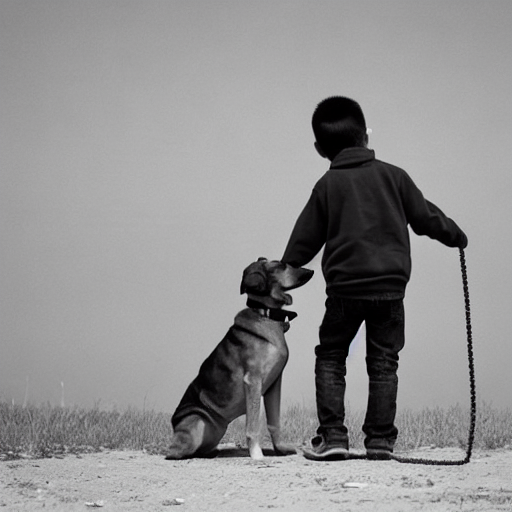} & \includegraphics[width=.34\linewidth,valign=m]{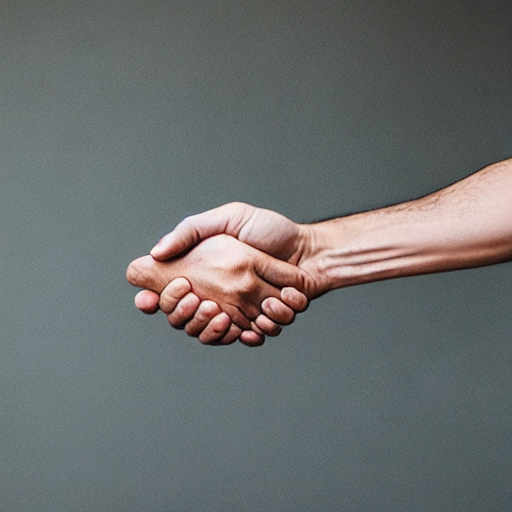} & \includegraphics[width=.34\linewidth,valign=m]{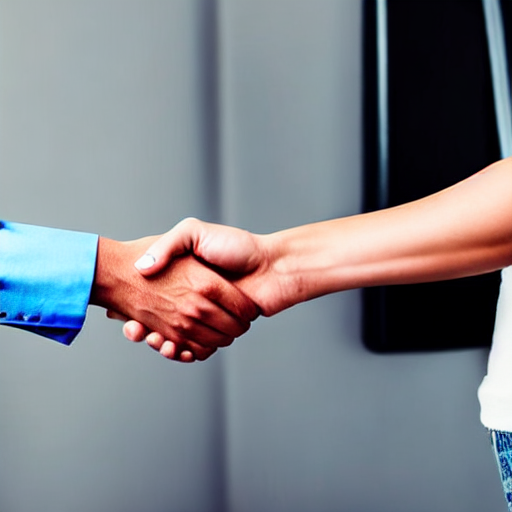}\\

\multicolumn{2}{c}{\color{darkgreen} ${\bf p}$: a man in a sombrero} & \multicolumn{2}{c}{\color{darkgreen} ${\bf p}$: a boy playing with his dog} & \multicolumn{2}{c}{ \color{darkgreen}${\bf p}$: a closeup of a handshake}\\
\multicolumn{2}{c}{ \color{darkred}${\bf n^*}$: a modern three story townhouse with two garages} & \multicolumn{2}{c}{ \color{darkred}${\bf n^*}$: three bowls of stew with vegetables and herbs} & \multicolumn{2}{c}{ \color{darkred}${\bf n^*}$: an aerial view of a large home in the woods}\\[0.1in]

\includegraphics[width=.34\linewidth,valign=m]{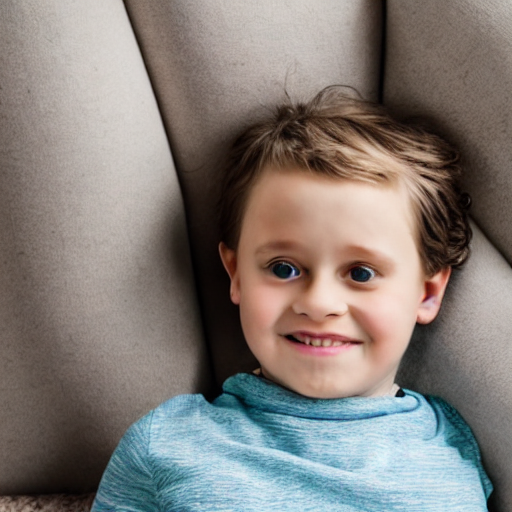} & \includegraphics[width=.34\linewidth,valign=m]{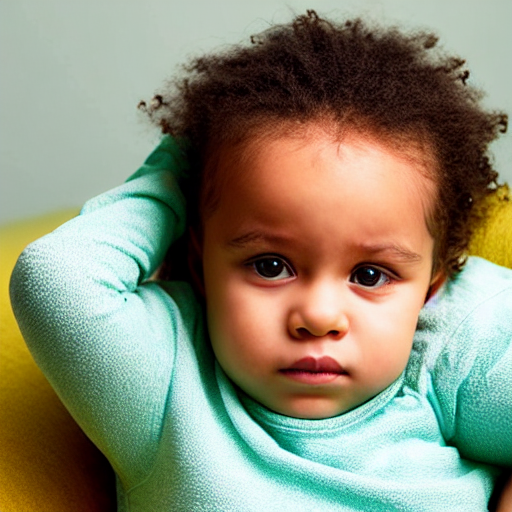} & \includegraphics[width=.34\linewidth,valign=m]{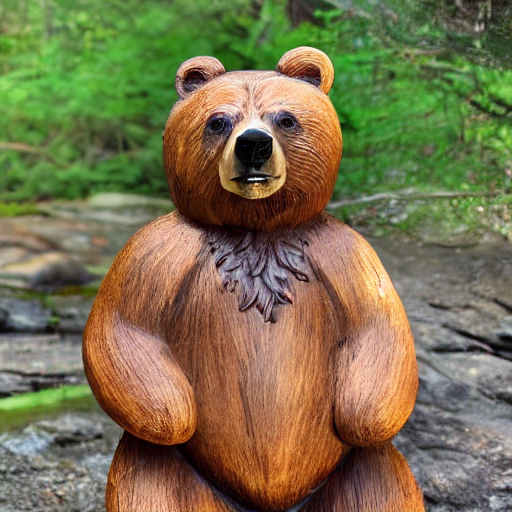} &
\includegraphics[width=.34\linewidth,valign=m]{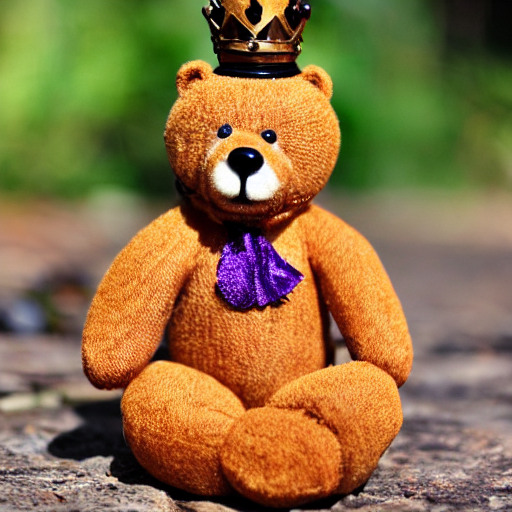} & \includegraphics[width=.34\linewidth,valign=m]{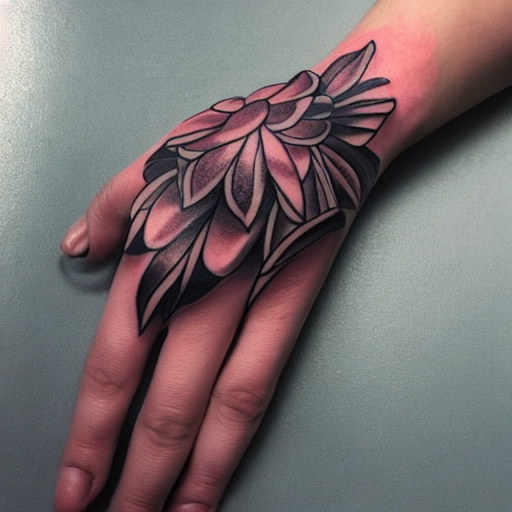} & \includegraphics[width=.34\linewidth,valign=m]{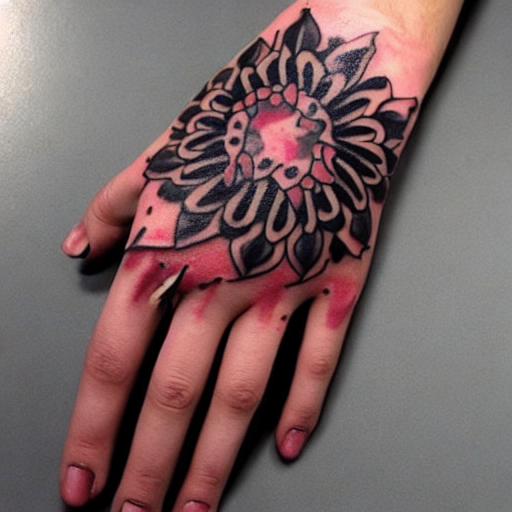}\\

\multicolumn{2}{c}{\color{darkgreen} ${\bf p}$: a child on the couch} & \multicolumn{2}{c}{ \color{darkgreen}${\bf p}$: a bear with a brown crown} & \multicolumn{2}{c}{\color{darkgreen} ${\bf p}$: a tattooed hand}\\
\multicolumn{2}{c}{ \color{darkred}${\bf n^*}$: men and women in suits and ties in front of a crowd} & \multicolumn{2}{c}{ \color{darkred}${\bf n^*}$: a rendering of the plans for a home} & \multicolumn{2}{c}{\color{darkred} ${\bf n^*}$: four pictures of a house, a car and a tree}\\

\end{tabular}}

\captionof{figure}{
    Images synthesized by SD (left) for prompt ${\bf p}$ vs. SD+\dna (right) for prompt pair (${\bf p}$, ${\bf n}^*$), where ${\bf n}^*$ is the \dnp estimated from the \dns image. \dnp produces negative prompts that are not intuitive for humans but improve quality and adherence.
    }
\label{fig:teaser}
\end{figure}

\subsection{Qualitative Results}

Figure~\ref{fig:teaser} shows qualitative examples from the three datasets, comparing SD with SD+\dna. There are more visual examples in \supp{Supplementary Section 5}.

The H\&H dataset results, e.g. the images synthesized for prompt ``\textit{hand with a ring on it}'', illustrate how \dna improves the quality of humans and hands in terms of the number of limbs, fingers, poses, etc.  It also induces the model to create realistic images, rather than drawings or sketches. See, for example, the ``\textit{man in a sombrero}'' and ``\textit{boy in the moonlight}'' examples. We observe a major improvement in the quality of the humans generated by \dna wherein the faces do not exhibit the uncanny, robotic, or distorted features that SD is inclined to create, instead having a natural warmth.

The results on the A\&E dataset show that \dna can correct for entity neglect while ensuring superior image quality. This is especially visible when we compare \textit{A\&E} with \textit{A\&E}+\dna (\supp{see Figures 5,6,7 in Supplementary Section 5}), which produces more realistic images. We posit that \dna's inclination towards generating more realistic images is due to setting the diffusion process onto a better Markov chain without disrupting it at each step, unlike \textit{A\&E}. 

Finally, note that none of the negative prompts produced by \dna in Figure~\ref{fig:teaser} is intuitive for humans. This illustrates the semantic gap between humans and SD and the difficulty of producing good negative prompts manually. 

\subsection{Semantic Gap}
\begin{figure}[t]
\centering
\resizebox{\linewidth}{!}{
\begin{tabular}{c@{\hskip 0.2em}c@{\hskip 0.2em}c@{\hskip 0.2em}c@{\hskip 0.3em} | @{\hskip 0.3em}c@{\hskip 0.2em}c@{\hskip 0.2em}c@{\hskip 0.2em}c}

\textbf{Stable} & \textbf{Diffusion-Negative} & \textbf{Stable} & \multirow{2}{*}{\textbf{SD+\dnp}} &
\textbf{Stable} & \textbf{Diffusion-Negative} & \textbf{Stable} & \multirow{2}{*}{\textbf{SD+\dnp}} \\

\textbf{Diffusion (SD)} & \textbf{Sampling} \textbf{(\dns)} & \textbf{Diffusion (SD)} &  & 
\textbf{Diffusion (SD)} & \textbf{Sampling} \textbf{(\dns)} & \textbf{Diffusion (SD)} &  \\

\includegraphics[width=0.25\linewidth]{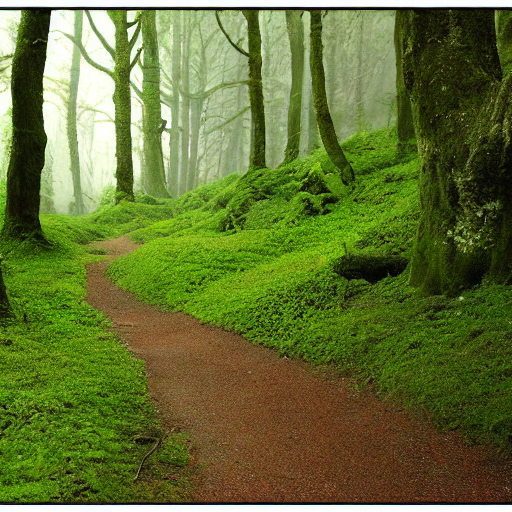}
&\includegraphics[width=0.25\linewidth]{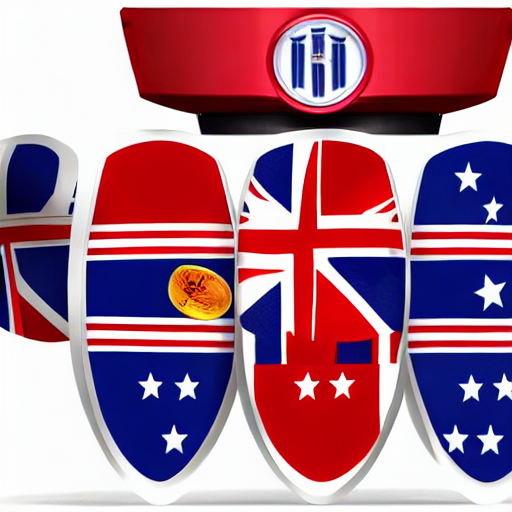}
&\includegraphics[width=0.25\linewidth]{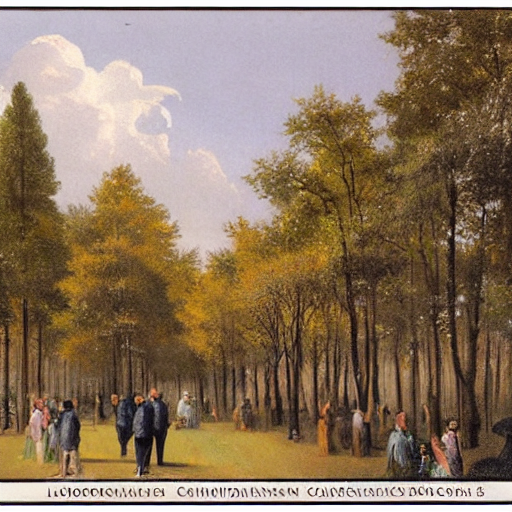}
&\includegraphics[width=0.25\linewidth]{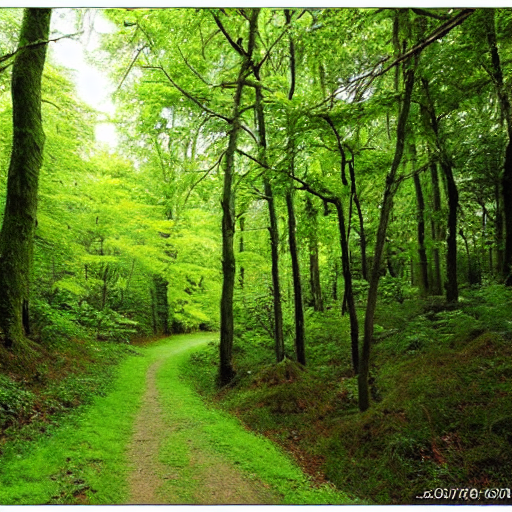} &

\includegraphics[width=0.25\linewidth]{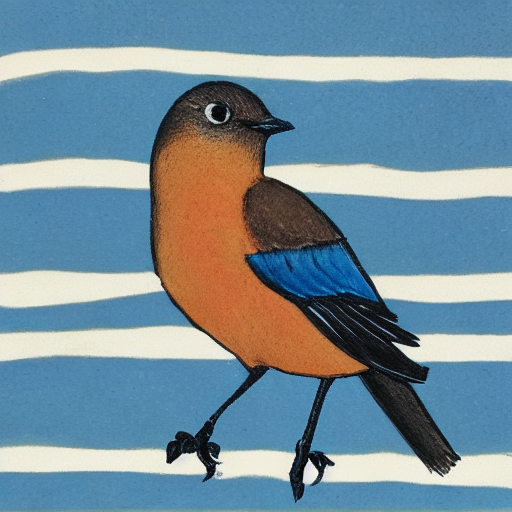}
&\includegraphics[width=0.25\linewidth]{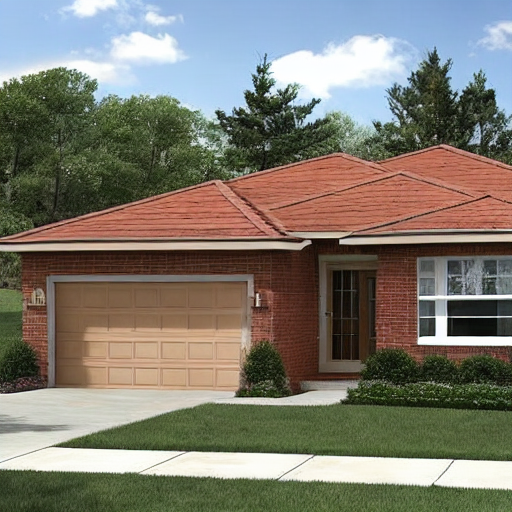}
&\includegraphics[width=0.25\linewidth]{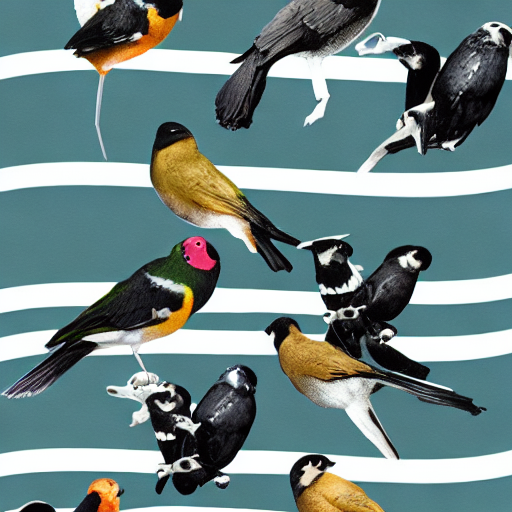}
&\includegraphics[width=0.25\linewidth]{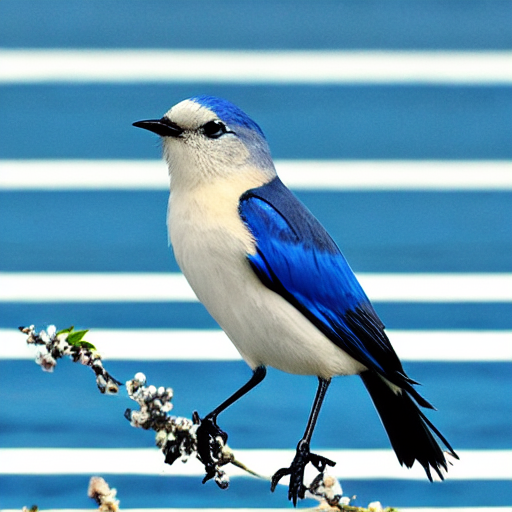}\\

{\color{darkgreen} ${\bf p}$: a path in a } & {${\bf {\bf n}^*}$: union jack }& \multirow{2}{*}{{\color{darkgreen} ${\bf p}$: ""}}& {\color{darkgreen} ${\bf p}$: a path in a } &
\multirow{2}{*}{{\color{darkgreen} ${\bf p}$: a blue bird}} & {${\bf {\bf n}^*}$: a red brick  }& \multirow{2}{*}{{\color{darkgreen} ${\bf p}$: ""}}& \multirow{2}{*}{{\color{darkgreen} ${\bf p}$: a blue bird}}\\

{\color{darkgreen} green forrest} & { on }&  & {\color{darkgreen} green forrest} &
 & { house }&  & \\
 
 {\color{darkred} ${\bf n}$: ""} & a surfboard &{\color{darkred} ${\bf n}$: ${{\bf n}^*}$} & {\color{darkred} ${\bf n}$: ${{\bf n}^*}$} &
 {\color{darkred} ${\bf n}$: ""} & with garage&{\color{darkred} ${\bf n}$: ${{\bf n}^*}$} & {\color{darkred} ${\bf n}$: ${{\bf n}^*}$}\\
 
\end{tabular}}
\captionof{figure}{\textbf{Exploring Semantic Gap:} From left to right: 1) images generated for prompt ${\bf p}$, 2) \dns images generated by the DM and resulting caption ${\bf n}^*$ by \dnp, 3) images synthesized with ${\bf n}^*$ as the negative prompt alone, and 4) with prompt pair (${\bf p}$, ${\bf n}^*$).}  
\label{fig:meth_exp}
\end{figure}

 Figure~\ref{fig:meth_exp} explores the semantic gap between humans and DMs in more detail. From left to right, the first column shows images synthesized by SD with prompt $\bf p$. The second column shows examples of images sampled with \dns, i.e. (\ref{eq:diff_inf_neg}). These can be seen as images that the model hallucinates as ``negatives'' of the prompt ${\bf p}$ since they usually do not have this interpretation for humans. The caption ${\bf n}^*$ is obtained the user. To confirm that the prompt mirrors the DM internal representation of the ``negative'' for the image in the left column, we prompted the model with an empty positive prompt ${\bf p}= {\bf \phi}$ and ${\bf n}^*$ as a negative prompt, which produces the images in the third column of the figure. Note how these images comply with $\bf p$ even though ${\bf n}^*$ is non-intuitive for a human. Finally, the last column shows the result of prompting with the pair ($\bf p$, $\bf n^*$). In all cases, the synthesized image is at least as good as that on the first column, and usually better, e.g. a greener forest or a more realistic bird.

Two aspects are worth noting. First, these examples purposefully uses prompts $\bf p$ for which SD works well. This proves that the ``DM knows what it's doing,'' at least to the point of generating a sensible image (first column). However, its notion of negative is totally different from that of humans, demonstrating the semantic gap between the two. The practical result is that a totally unintuitive negative prompt is needed to produce good images. Figure~\ref{fig:teaser} shows that \dna can improve the quality and correctness of the synthesised image even when the DM fails. Second, because all derivations above are functions of $z_t$, the \dns procedure is valid for any seed $z_T$. This is an additional advantage over the human-centric negative prompting of Figure~\ref{fig:semantic_neg_example}, whose success is seed-dependent.

\section{Related Work}
\label{sec:relwork}
\noindent{\bf Text-to-Image Diffusion:}
High-resolution T2I generation has advanced dramatically with the introduction of large-scale diffusion models, due to: (1) availability of large-scale text-image datasets~\cite{lion5b,coyo700m}, (2) advances in various training and inference techniques~\cite{ddpm, ddim, cfg, cascaddiff}, and, (3) development of scalable model architectures~\cite{imagen, dalle, dalle2, ldm}. The most popular T2I models employ classifier-free guidance~\cite{cfg} to balance prompt compliance and image diversity using a guidance scale hyper-parameter, $s$. While prompt adherence improves by increasing $s$, image quality deteriorates beyond a certain value. \dnp helps ensure prompt adherence even for lower values of $s$, thereby maintaining image quality.

\noindent{\bf Structured Image Conditioning:} Various image generation and editing improvements have been introduced since the advent of DMs. They range from finetuning the models for certain tasks to editing the noise at each iteration. Many methods require users to provide structured conditioned inputs like layouts~\cite{gligen, groundeddiff,layoutdiff}, example images~\cite{kumari2023multi, dreambooth, pplus}, and depth maps~\cite{controlnet}. These can be hard to produce and require considerable user skill or additional computer vision modules. \dnp allows us to control T2I with text alone while allowing it to be combined with most structured conditioning techniques as well.  

\noindent\textbf{Attention-based Methods:} These methods attempt to tackle DM errors by editing attention maps at each diffusion step, changing the cross-attention between prompt tokens. \cite{ae} updates the noise latent with a loss that maximizes attention to each noun, while \cite{syngen} first creates a linguistic binding between prompt attributes and nouns and then maximizes (minimizes) attention IOUs for mapped (non-mapped) tokens. \cite{richtext} uses text formatting such as font style, size, color, and footnotes to allow for creative human input and uses region-specific guidance to combine the noise corresponding to each token. These methods require careful fine-tuning of a loss guidance hyper-parameter and can heavily alter the DM Markov chain, which can create highly saturated and low-quality images. While the proposed method also forces the DM to use a different Markov chain, it does not interfere with its intermediate steps through any loss updates.

\noindent\textbf{Text based Methods:}
These methods try to explore and correct the semantic failures of the DM text encoder. \cite{compos} tries to decouple the text and combine noise latent later and \cite{structdiff} introduces consistency trees to split the prompt into noun phrases, for better cross-attention in the text-encoder. These methods fail to correct those errors which arise not from the text but from the DM itself. 

\noindent\textbf{Negative Prompts:} Negative prompts~\cite{compos}, are quite effective at removing unwanted concepts from the synthesized image and have been incorporated into most T2I models. However, finding a good negative prompt requires trial and error. Also, the effect of negative prompts varies significantly with prompt and seed. The proposed method, \dnp provides a solution to this.

\section{Conclusion}

Do diffusion models truly understand negation? In this work, we have hypothesized that the semantic gap between DMs and humans is responsible for the poor performance of negative prompting. We proposed \dnp as a simple yet effective method for bridging this gap. This consists of sampling a negative diffusion image \dni, using a novel \dns procedure, and asking the user to translate it into a natural language negative prompt. This greatly improves the prompt adherence and quality of the generated images. \dnp is universally applicable across diffusion models and can be combined with other methods. It highlights the difference between semantic and diffusion negation and leverages this difference to improve the performance of these models without additional training.  

\section*{Acknowledgements}
This work was partially funded by NSF award IIS-2303153 a gift from Qualcomm, and NVIDIA GPU donations. We also acknowledge and thank the use of the Nautilus platform for the experiments discussed above.
%
%
\bibliographystyle{splncs04}
\bibliography{bibliography}

\begin{thebibliography}{10}
\providecommand{\url}[1]{\texttt{#1}}
\providecommand{\urlprefix}{URL }
\providecommand{\doi}[1]{https://doi.org/#1}

\bibitem{AUTOMATIC1111}
AUTOMATIC1111: Negative prompt: Stable diffusion webui, \url{https://github.com/AUTOMATIC1111/stable-diffusion-webui/wiki/Negative-prompt}

\bibitem{coyo700m}
Byeon, M., Park, B., Kim, H., Lee, S., Baek, W., Kim, S.: Coyo-700m: Image-text pair dataset. \url{https://github.com/kakaobrain/coyo-dataset} (2022)

\bibitem{ae}
Chefer, H., Alaluf, Y., Vinker, Y., Wolf, L., Cohen-Or, D.: Attend-and-excite: Attention-based semantic guidance for text-to-image diffusion models (2023)

\bibitem{improperdist}
Dawid, A.P., Stone, M., Zidek, J.V.: Marginalization paradoxes in bayesian and structural inference. Journal of the Royal Statistical Society Series B: Statistical Methodology  \textbf{35}(2),  189--213 (1973)

\bibitem{compos_prev}
Du, Y., Li, S., Mordatch, I.: Compositional visual generation with energy based models. In: Larochelle, H., Ranzato, M., Hadsell, R., Balcan, M., Lin, H. (eds.) Advances in Neural Information Processing Systems. vol.~33, pp. 6637--6647. Curran Associates, Inc. (2020), \url{https://proceedings.neurips.cc/paper_files/paper/2020/file/49856ed476ad01fcff881d57e161d73f-Paper.pdf}

\bibitem{structdiff}
Feng, W., He, X., Fu, T.J., Jampani, V., Akula, A., Narayana, P., Basu, S., Wang, X.E., Wang, W.Y.: Training-free structured diffusion guidance for compositional text-to-image synthesis (2023)

\bibitem{richtext}
Ge, S., Park, T., Zhu, J.Y., Huang, J.B.: Expressive text-to-image generation with rich text. In: IEEE International Conference on Computer Vision (ICCV) (2023)

\bibitem{clipscore}
Hessel, J., Holtzman, A., Forbes, M., Bras, R.L., Choi, Y.: Clipscore: A reference-free evaluation metric for image captioning (2022)

\bibitem{ddpm}
Ho, J., Jain, A., Abbeel, P.: Denoising diffusion probabilistic models. Advances in neural information processing systems  \textbf{33},  6840--6851 (2020)

\bibitem{cascaddiff}
Ho, J., Saharia, C., Chan, W., Fleet, D.J., Norouzi, M., Salimans, T.: Cascaded diffusion models for high fidelity image generation. The Journal of Machine Learning Research  \textbf{23}(1),  2249--2281 (2022)

\bibitem{cfg}
Ho, J., Salimans, T.: Classifier-free diffusion guidance. arXiv preprint arXiv:2207.12598  (2022)

\bibitem{text2human}
Jiang, Y., Yang, S., Qiu, H., Wu, W., Loy, C.C., Liu, Z.: Text2human: Text-driven controllable human image generation (2022)

\bibitem{kumari2023multi}
Kumari, N., Zhang, B., Zhang, R., Shechtman, E., Zhu, J.Y.: Multi-concept customization of text-to-image diffusion. In: Proceedings of the IEEE/CVF Conference on Computer Vision and Pattern Recognition. pp. 1931--1941 (2023)

\bibitem{blip2}
Li, J., Li, D., Savarese, S., Hoi, S.: Blip-2: Bootstrapping language-image pre-training with frozen image encoders and large language models (2023)

\bibitem{gligen}
Li, Y., Liu, H., Wu, Q., Mu, F., Yang, J., Gao, J., Li, C., Lee, Y.J.: Gligen: Open-set grounded text-to-image generation. In: Proceedings of the IEEE/CVF Conference on Computer Vision and Pattern Recognition. pp. 22511--22521 (2023)

\bibitem{compos}
Liu, N., Li, S., Du, Y., Torralba, A., Tenenbaum, J.B.: Compositional visual generation with composable diffusion models. In: European Conference on Computer Vision. pp. 423--439. Springer (2022)

\bibitem{text2hand}
Lu, W., Xu, Y., Zhang, J., Wang, C., Tao, D.: Handrefiner: Refining malformed hands in generated images by diffusion-based conditional inpainting (2023)

\bibitem{groundeddiff}
Phung, Q., Ge, S., Huang, J.B.: Grounded text-to-image synthesis with attention refocusing. arXiv preprint arXiv:2306.05427  (2023)

\bibitem{dalle2}
Ramesh, A., Dhariwal, P., Nichol, A., Chu, C., Chen, M.: Hierarchical text-conditional image generation with clip latents. arXiv preprint arXiv:2204.06125  \textbf{1}(2), ~3 (2022)

\bibitem{dalle}
Ramesh, A., Pavlov, M., Goh, G., Gray, S., Voss, C., Radford, A., Chen, M., Sutskever, I.: Zero-shot text-to-image generation. In: International Conference on Machine Learning. pp. 8821--8831. PMLR (2021)

\bibitem{syngen}
Rassin, R., Hirsch, E., Glickman, D., Ravfogel, S., Goldberg, Y., Chechik, G.: Linguistic binding in diffusion models: Enhancing attribute correspondence through attention map alignment (2023)

\bibitem{ldm}
Rombach, R., Blattmann, A., Lorenz, D., Esser, P., Ommer, B.: High-resolution image synthesis with latent diffusion models. In: Proceedings of the IEEE/CVF conference on computer vision and pattern recognition. pp. 10684--10695 (2022)

\bibitem{dreambooth}
Ruiz, N., Li, Y., Jampani, V., Pritch, Y., Rubinstein, M., Aberman, K.: Dreambooth: Fine tuning text-to-image diffusion models for subject-driven generation. In: Proceedings of the IEEE/CVF Conference on Computer Vision and Pattern Recognition (CVPR). pp. 22500--22510 (June 2023)

\bibitem{imagen}
Saharia, C., Chan, W., Saxena, S., Li, L., Whang, J., Denton, E.L., Ghasemipour, K., Gontijo~Lopes, R., Karagol~Ayan, B., Salimans, T., et~al.: Photorealistic text-to-image diffusion models with deep language understanding. Advances in Neural Information Processing Systems  \textbf{35},  36479--36494 (2022)

\bibitem{IS}
Salimans, T., Goodfellow, I., Zaremba, W., Cheung, V., Radford, A., Chen, X.: Improved techniques for training gans (2016)

\bibitem{lion5b}
Schuhmann, C., Beaumont, R., Vencu, R., Gordon, C., Wightman, R., Cherti, M., Coombes, T., Katta, A., Mullis, C., Wortsman, M., Schramowski, P., Kundurthy, S., Crowson, K., Schmidt, L., Kaczmarczyk, R., Jitsev, J.: Laion-5b: An open large-scale dataset for training next generation image-text models. In: Koyejo, S., Mohamed, S., Agarwal, A., Belgrave, D., Cho, K., Oh, A. (eds.) Advances in Neural Information Processing Systems. vol.~35, pp. 25278--25294. Curran Associates, Inc. (2022), \url{https://proceedings.neurips.cc/paper_files/paper/2022/file/a1859debfb3b59d094f3504d5ebb6c25-Paper-Datasets_and_Benchmarks.pdf}

\bibitem{ddim}
Song, J., Meng, C., Ermon, S.: Denoising diffusion implicit models. arXiv preprint arXiv:2010.02502  (2020)

\bibitem{song2019generative}
Song, Y., Ermon, S.: Generative modeling by estimating gradients of the data distribution. Advances in neural information processing systems  \textbf{32} (2019)

\bibitem{pplus}
Voynov, A., Chu, Q., Cohen-Or, D., Aberman, K.: P+: Extended textual conditioning in text-to-image generation (2023)

\bibitem{longevindyn}
Welling, M., Teh, Y.W.: Bayesian learning via stochastic gradient langevin dynamics. In: Proceedings of the 28th international conference on machine learning (ICML-11). pp. 681--688 (2011)

\bibitem{controlnet}
Zhang, L., Rao, A., Agrawala, M.: Adding conditional control to text-to-image diffusion models (2023)

\bibitem{layoutdiff}
Zheng, G., Zhou, X., Li, X., Qi, Z., Shan, Y., Li, X.: Layoutdiffusion: Controllable diffusion model for layout-to-image generation. In: Proceedings of the IEEE/CVF Conference on Computer Vision and Pattern Recognition. pp. 22490--22499 (2023)

\end{thebibliography}

\end{document}